\documentclass[letterpaper]{article} 
\usepackage{aaai25}
\usepackage{times}  
\usepackage{helvet}  
\usepackage{courier}  
\usepackage[hyphens]{url}  
\usepackage{graphicx} 
\usepackage{natbib}  
\usepackage{caption} 
\usepackage{fancyhdr} 
\fancypagestyle{firstpage}{
    \fancyhead[L]{} 
    \fancyhead[C]{The Thirty-Ninth AAAI Conference on Artificial Intelligence (AAAI-25)} 
    \fancyhead[R]{} 
}

\usepackage{algorithm}
\usepackage{algorithmic}
\usepackage{amsmath}
\usepackage{amssymb}
\usepackage{booktabs}
\usepackage{color}
\usepackage{multirow}

\def\code{\textcolor[rgb]{0.98, 0.28, 0.2}}
\def\xie{\textcolor{black}}
\def\wang{\textcolor{black}}
\def\leng{\textcolor{black}}
\def\todo{\textcolor{black}}
\usepackage{newfloat}
\usepackage{listings}
\DeclareCaptionStyle{ruled}{labelfont=normalfont,labelsep=colon,strut=off} 
\floatstyle{ruled}
\newfloat{listing}{tb}{lst}{}
\floatname{listing}{Listing}
\title{IAA: Inner-Adaptor Architecture Empowers Frozen Large Language Model with Multimodal Capabilities}
\author{
    Bin Wang\equalcontrib,   
    Chunyu Xie\equalcontrib,
    Dawei Leng\thanks{Corresponding author.},
    Yuhui Yin  
}
\affiliations{
    360 AI Research \\
    \{wangbin10, xiechunyu, lengdawei, yinyuhui\}@360.cn \\

}

\usepackage{bibentry}

\begin{document}

\maketitle

\begin{abstract}

\xie{In the field of multimodal large language models (MLLMs), common methods typically involve unfreezing the language model during training to foster profound visual understanding. However, the fine-tuning of such models with vision-language data often leads to a diminution of their natural language processing (NLP) capabilities. To avoid this performance degradation, a straightforward solution is to freeze the language model while developing multimodal competencies. Unfortunately, previous works have not attained satisfactory outcomes. Building on the strategy of freezing the language model, we conduct thorough structural exploration and introduce the Inner-Adaptor Architecture (IAA). Specifically, the architecture incorporates multiple multimodal adaptors at varying depths within the large language model to facilitate direct interaction with the inherently text-oriented transformer layers, thereby enabling the frozen language model to acquire multimodal capabilities. Unlike previous approaches of freezing language models that require large-scale aligned data, our proposed architecture is able to achieve superior performance on small-scale datasets. We conduct extensive experiments to improve the general multimodal capabilities and visual grounding abilities of the MLLM. Our approach remarkably outperforms previous state-of-the-art methods across various vision-language benchmarks without sacrificing performance on NLP tasks. Code and models are available at \code{https://github.com/360CVGroup/Inner-Adaptor-Architecture}.}
\end{abstract}

%

\section{Introduction}

\xie{Large Language Models (LLMs) have made substantial progress in recent years, largely attributed to the technique of pre-training and instruction tuning. Building upon this foundation, visual instruction tuning has been proposed to evolve LLMs into Multimodal Large Language Models (MLLMs), thereby endowing them with the capability to interpret and comprehend visual signals \cite{cha2024honeybee}.
MLLMs~\cite{llava,bai2023qwen,tong2024cambrian,chen2024internvl,xuan2024pink} prove beneficial in numerous tasks, \leng{such as transcribing the text within an image, generating stories and poems based on an image, or converting screenshots of webpages into code \cite{idefics}}. Historically, these tasks have been regarded as challenging for conventional vision-language models. MLLMs exhibit considerable promise in executing these complex, diverse real-world tasks, enabling more natural and human-like interactions~\cite{lu2024deepseek}.}

\xie{Typically, the operation of a MLLM begins with feeding an image into a visual encoder, such as CLIP \cite{radford2021learning} or SigLIP \cite{zhai2023sigmoid}, to extract a high-dimensional feature representation. This feature is subsequently transformed through a projection layer to align with the dimension of the large language model. The resulting features, often referred to as image tokens, are concatenated with text tokens and fed into the large language model. This process enables the MLLM to generate responses based on user instructions and input images.}

\begin{figure}[!tbp]
  \centering
   \includegraphics[width=0.99\columnwidth]{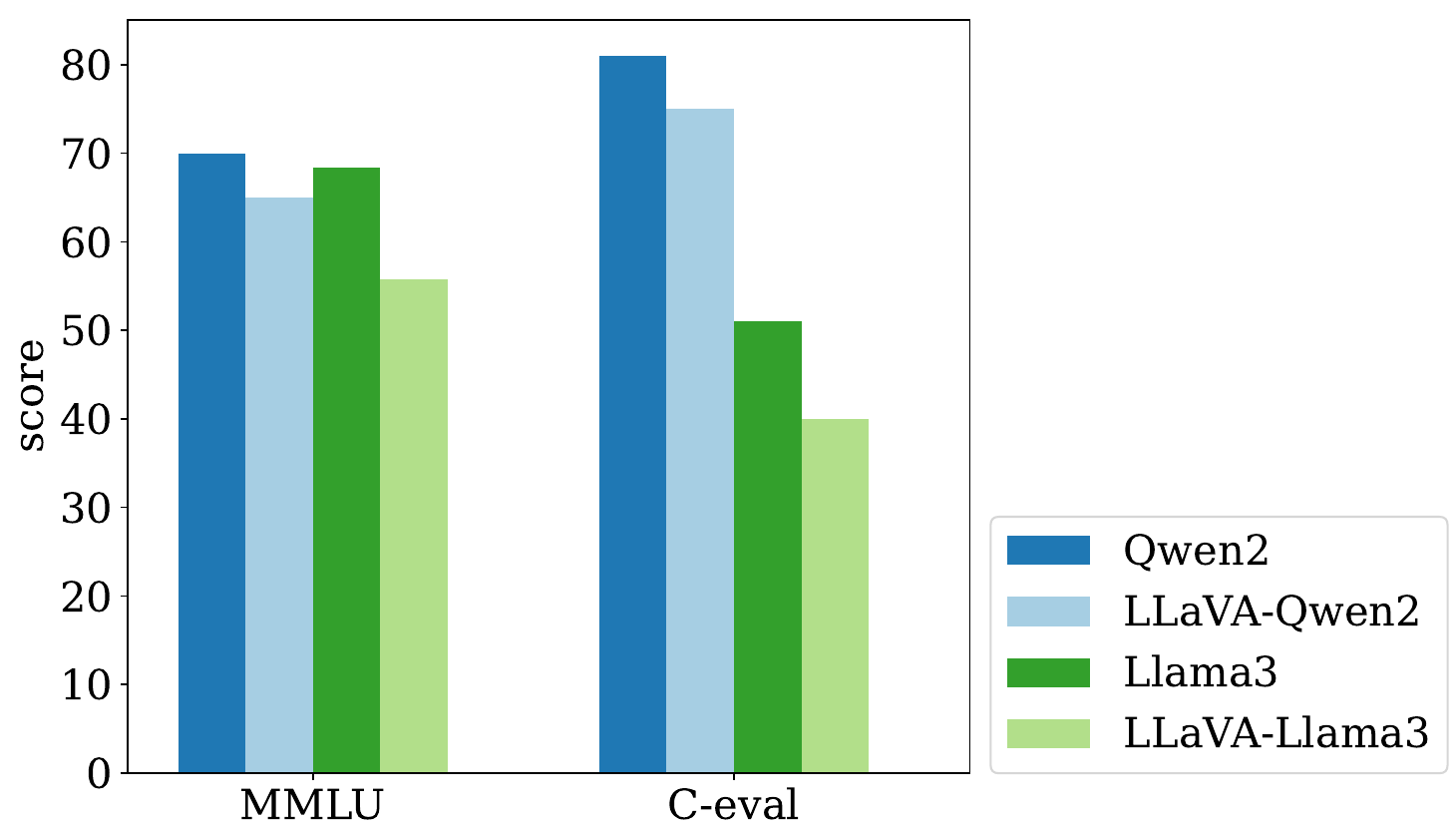}
   \caption{Results before and after training LLaVA-1.5 architecture based on Qwen2 and Llama3 language models on text-only evaluation set MMLU and C-eval.}
    \label{fig:barchart}
\end{figure}

\xie{In the current common MLLM \cite{liu2024improved,bai2023qwen}, when image and text tokens are fed into the large language model, the LLM is typically unfrozen for further training. \leng{This strategy has led to significant advancements in the MLLM model.} Consequently, it predictably leads to a degradation in the understanding ability of the large language model. To validate this hypothesis, we conduct experiments on the LLaVA-1.5 \cite{liu2024improved} architecture using the 1.2M-size open-source dataset provided by \cite{liu2024improved}, which contains a limited amount of plain text data, as illustrated in Figure \ref{fig:barchart}. We compare the results before and after training the LLaVA-1.5 architecture, based on the Qwen2 \cite{qwen2} and Llama3 \cite{llama3} language models, respectively. The performance of the language model declines significantly on both the MMLU \cite{hendrycks2020measuring} and C-Eval \cite{huang2023ceval} text-only evaluation sets.}

\xie{It appears reasonable to posit an explanation for this phenomenon within the field of deep learning. When a model is predominantly trained on a single type of data, it may experience a phenomenon known as catastrophic forgetting. For an MLLM to achieve outstanding image-text comprehension, it is essential to collect a substantial amount of image-text interaction data for training. As observed in Figure \ref{fig:barchart}, training with image-text data results in a decline in language ability. Despite attempts by MLLM such as LLaVA to incorporate some text-only data into their training process, this still leads to a reduction in the model's comprehension. }

\xie{One direct method to prevent the degradation of LLM performance is to freeze the large language model during the training of MLLM. However, current methods employing this approach \cite{li2023blip,zhu2023minigpt} have consistently struggled to achieve powerful multimodal capabilities. To address these challenges, we propose a new training paradigm with an inner-adaptor architecture that significantly enhances multimodal competencies without affecting the original language modeling capabilities.
This approach can seamlessly support both multimodal and textual workflows. We evaluate this training paradigm across a spectrum of tasks, including general multimodal capabilities and visual grounding proficiencies. Distinct from previous approaches of freezing language modeling that require large-scale aligned data, our proposed scheme demonstrates effectiveness with a considerably smaller dataset. }
\leng{Comprehensive testing on a suite of benchmarks, including MME, MMBench, MMMU, and RefCOCO, has substantiated the superior performance of our structure. We hope that this approach will provide a reference for future research in open-source MLLM.}

\section{Related Work}
\subsubsection{Large Language Models.} 
\xie{The landscape of Natural Language Processing (NLP) has undergone a revolutionary transformation, driven by the advent and continuous refinement of Large Language Models (LLMs). A pivotal moment in this evolution is \leng{the first appearance of the transformer architecture}, which serves as a key catalyst, giving rise to pioneering language models like BERT \cite{bert} and OPT \cite{opt}. These models showcase an unprecedented level of linguistic comprehension, significantly advancing the state-of-the-art in NLP. A critical breakthrough comes with introducing the Generative Pre-trained Transformer (GPT) series \cite{gpt}, which pioneer an auto-regressive language modeling approach, setting a new standard for language prediction and generation capabilities. Subsequent iterations, including Mixtral \cite{mixtral}, GPT-4 \cite{gpt4}, and Llama3 \cite{llama3}, have not only maintained but also amplified this momentum, displaying superior performance on intricate language processing challenges. Moreover, the fusion of LLMs with specialized visual tasks showcases the models' adaptability and broadens their scope, indicating their potential to transcend conventional text-based operations into multimodal interactions. This expansion highlights the transformative role LLMs can assume when incorporated into diverse domains, providing a rich ground for innovation and exploration.}

\subsubsection{Multimodal Large Language Models.}
\xie{The advancement of Large Language Models (LLMs) has kindled a growing interest in extending their foundational competencies to incorporate the visual domain, thereby giving birth to multimodal Large Language Models (MLLMs).
The works on MLLMs~\cite{xie2023ccmb,li2023makes,li2023blip,bai2023qwen,llava,idefics,chen2024internvl} typically follow a tripartite architecture: a visual encoder, a vision-language connector, and a large language model. Notably, BLIP-2~\cite{li2023blip} and Flamingo~\cite{flamingo} introduce the Q-Former/Resampler as a bridge between vision and language, whereas LLaVA~\cite{llava} and MiniGPT4~\cite{zhu2023minigpt} refine this connection via a linear layer. Cambrian-1~\cite{tong2024cambrian} proposes a dynamically adaptive connector that integrates high-resolution visual features with LLMs while reducing the number of tokens. To enhance their multimodal performance, contemporary MLLMs mainly fine-tune the LLM and connector using visual instruction tuning data. These models leverage meticulously curated instruction datasets, showcasing an effective strategy that highlights their robust capabilities. However, a common oversight lies in the maintenance of language abilities. \leng{Long term multimodal training often leads to degradation of language proficiency.}
CogVLM~\cite{wang2023cogvlm} seeks to address this by integrating a trainable visual expert into the language model, but still trains the LLM during supervised fine-tuning, resulting in a degradation of language capability. DeekSeek-VL~\cite{lu2024deepseek} maintains a 70\% proportion of language data to preserve the integrity of language knowledge within the model, but incurs a considerable training cost. Departing from these conventional training paradigms of MLLMs, we introduce the inner-adaptor architecture. This design is specifically tailored to preserve the NLP performance of the MLLM while facilitating a seamless augmentation of its multimodal capabilities.}

\begin{figure*}[!htbp]
  \centering   \includegraphics[width=1.0\linewidth]{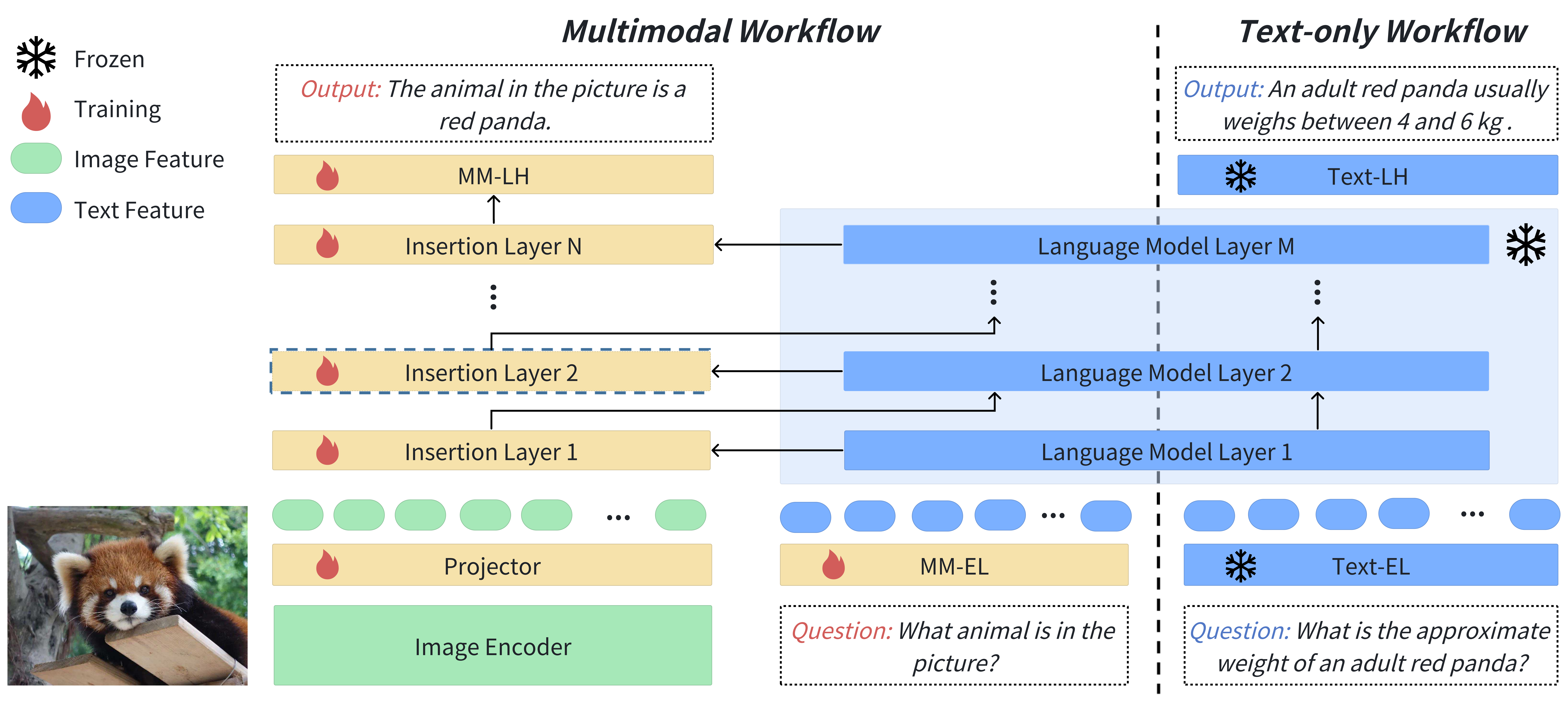}
   
    \caption{\xie{Overview of the proposed architecture, which mainly consists of two workflows: the Multimodal Workflow and the Text-only Workflow. The multimodal workflow, beyond the necessary image encoder and projector, integrates the Inner-Adaptor Architecture, including insertion layers, an embedding layer, and a language model head. Both workflows share the same large language model. The number of insertion layers is variable, where $N \leq M $. In this context, \textit{MM} denotes MultiModal, \textit{EL} stands for Embedding Layer, and \textit{LH} represents the Language model Head.}}
    
    \label{fig:arch}
\end{figure*}

\section{Methodology}

\subsubsection{Overview.}
\xie{As illustrated in Figure \ref{fig:arch}, our approach enables the simultaneous execution of two high-quality workflows post-deployment: one for multimodal interactions and the other for text-only conversations. Both workflows leverage the transformer layers of the large language model. The multimodal interaction workflow encompasses: (1) an image encoder and a projector, utilized for extracting high-quality image features and achieving vision-language alignment, respectively, (2) the transformer layers of the large language model, which remain frozen during training, and (3) the inner-adaptor architecture, which comprises insertion layers, an embedding layer, and a language model head specifically designed for multimodal inputs.
Conversely, the text-only conversation workflow solely employs the constituent elements of the original language model, without resorting to the specialized multimodal components.}

\subsubsection{Image Encoder and Projector.}
Following LLava-1.5 \cite{liu2024improved}, we utilize the CLIP ViT-L/14 \cite{radford2021learning} image encoder with an input resolution of 336px. Subsequently, we employ a vision-language projector composed of a two-layer MLP to integrate the vision features with LLMs.

\subsubsection{Large Language Model.}
We employ the Llama3-8B \cite{llama3} as the base language model throughout the training process. 

\subsubsection{Inner-Adaptor Architecture.}To achieve multimodal comprehension, it is essential to integrate trainable parameters into MLLMs. 
\xie{LLaVA~\cite{llava} makes the projector and the large language model trainable during visual instruction tuning, but leads to the performance degradation on NLP tasks. Flamingo \cite{flamingo} employs cross-attention with a gating mechanism to introduce image information into the model, facilitating a deep fusion of original image features with text features prior to each layer of the language model. However, this approach requires a considerable volume of pre-training data to train effective cross-attention layers and gating values, which can be computationally costly. Furthermore, the final performance of the model falls short of expectations.}


\begin{figure*}[!htbp]
  \centering
   \includegraphics[width=1\linewidth]{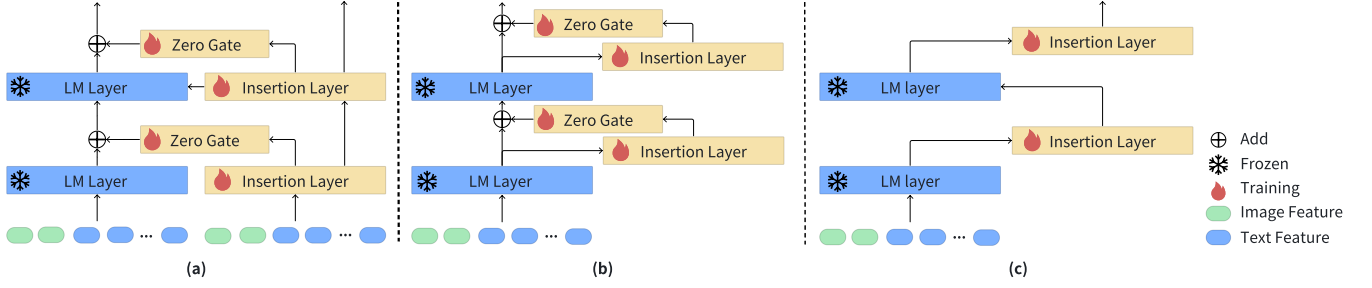}
   \caption{\xie{Structural exploration of the Inner-Adaptor Architecture. Figure (a) is a architecture inspired by the ControlNet design; Figure (b) is an improvement on Figure (a), mainly canceling the feature propagation between adaptors; Figure (c) is the final scheme.}}
    \label{fig:subarch}
\end{figure*}

Drawing insights from recent works~\cite{chen2024allava,tong2024cambrian}, we recognize that the self-attention layer can assimilate image features as prior prompts, thus eliminating the necessity of cross-attention for the obligatory incorporation of image features. In alignment with this perspective, we embark on exploratory research. Referencing Figure \ref{fig:subarch}(a), we are inspired by the prevalent ControlNet~\cite{controlnet} architecture. The operation of a specific layer can be succinctly expressed as follows:
\begin{align}\label{eq:control}
 \footnotesize
    &X_{out}=\phi_{fl}(X_{in})+G(\phi_{il}(X_{in})),
    \end{align}
where $\phi_{fl}$ and $\phi_{il}$ denote the frozen language model (LM) layer and the insertion layer, respectively.
Here, $X_{in}$ represents the multimodal input, $X_{out}$ denotes the multimodal output, and $G$ indicates a gating layer initialized at zero. 
\leng{The insertion layer is a transformer decoder layer, comprising the self-attention layer, layer normalization, feed forward network, etc. It is consistent with the parameter scale of a transformer layer in the large language model.} For instance, if we target the $22th$ layer, the initial parameters of the corresponding insertion layer are derived from the $22th$ language model layer. Nonetheless, the ControlNet-based design did not yield satisfactory performance.

Referring to Figure \ref{fig:subarch}(b), we endeavor to refine the ControlNet structure. Specifically, we eliminate the feature propagation between insertion layers. Instead, the output of the LM layer serves as the input to the insertion layer. Our expectation is that each frozen LM layer will accommodate multimodal data through a distinct insertion layer and gating layer, with the insertion layer no longer being directly influenced by subsequent layers. Compared to the design in Figure \ref{fig:subarch}(a), the refined architecture shows significant improvements.

Moreover, we hypothesize that the gating layer may not reach an optimal state through a single round of data training. Consequently, we propose a more streamlined solution, as illustrated in Figure \ref{fig:subarch}(c). The operation of a specific layer within the model can be represented as follows:
 \begin{align}\label{eq:insert}
 \footnotesize
    &X_{out} = \phi_{il}(\phi_{fl}(X_{in})).
    \end{align}
Similar to Scheme (a), if an insertion layer is placed after the $22th$ LM layer, it is initialized from the parameters of the $22th$ frozen LM layer. The number of insertion layers is adjustable.

Additionally, for multimodal training, we introduce a new embedding layer $EL_{mm}$ and a new LM head $LH_{mm}$, initialized from the original language model's embedding layer $EL_{text}$ and LM head $LH_{text}$. Throughout all stages of multimodal training, $EL_{text}$ and $LH_{text}$ will remain frozen, while the newly created components will be trained with multimodal data. The experimental results presented in Table \ref{tab:aba_strc} validate the effectiveness of this strategy. 

We thoroughly explore the distinctions among these architectures and strategies in the ablation study. Ultimately, we select the structure depicted in Figure \ref{fig:subarch}(c), which we designate as the Inner-Adaptor Architecture (IAA).

\begin{table}[t]
      \centering
        \footnotesize  
      \label{fps_cost}
        \begin{tabular}{c|cc}
        \toprule
        Configurations &  Satge1-PT  & Satge2-PT  \\
        \midrule
        Trainable modules & Projector & \xie{Projector, Inner-adaptor }\\
        Learning rate&1e-3&2e-5\\
               Batch size&\multicolumn{2}{c}{256}\\
        LR schedule &\multicolumn{2}{c}{Cosine decay}\\
        Training steps &\multicolumn{2}{c}{2.5K}\\
        Zero-Stage& \multicolumn{2}{c}{Zero2}\\
        Warmup ratio &\multicolumn{2}{c}{0.03}\\
        Weight decay &\multicolumn{2}{c}{0.0}\\
        Optimizer &\multicolumn{2}{c}{AdamW}\\
        Optimizer HPs  &\multicolumn{2}{c}{$\beta_{1}=0.9$, $\beta_{2}=0.98$, $\epsilon=1e-6$}\\
        \midrule
                Configurations &  Instruction-FT  & Grounding-FT  \\
        \midrule
        Trainable modules &     \multicolumn{2}{c}{\xie{Projector, Inner-adaptor}}\\
                Learning rate&\multicolumn{2}{c}{2e-5}\\
        Batch size&\multicolumn{2}{c}{128}\\
        LR schedule & \multicolumn{2}{c}{Cosine decay}\\
        Training steps &6.6K&18K\\
         Zero-Stage& \multicolumn{2}{c}{Zero3}\\
        Warmup ratio &\multicolumn{2}{c}{0.03}\\
        Weight decay &\multicolumn{2}{c}{0.0}\\
        Optimizer &\multicolumn{2}{c}{AdamW}\\
        Optimizer HPs  &\multicolumn{2}{c}{$\beta_{1}=0.9$, $\beta_{2}=0.98$, $\epsilon=1e-6$}\\
        \bottomrule 
        \end{tabular}
    \caption{\xie{
    The hyperparameters utilized during the training phase are delineated as follows: "-PT" designates the pre-training phase, "-FT" denotes the fine-tuning phase, and "HP" and "LR" signify the hyperparameter and learning rate, respectively.}}
    \label{tab:hpmsg}
\end{table}

\section{Experiments}
\leng{In this section, we first describe the training paradigm of our method with the data utilized in the diverse processes. Subsequently, we conduct evaluation on the general multimodal and visual grounding benchmarks to comprehensively assess our models’ visual understanding ability. Finally, we detail the ablation experiments of our method.}
\subsection{Training Paradigm}
\subsubsection{Pre-training.}

\xie{During the training process of MLLM, the primary objective of the pre-training phase is to enable MLLM to learn the alignment between visual cues and textual descriptions. This stage, also known as the image-text alignment phase, establishes connections between the vision encoder and LLM. In our architectural design, the image encoder and LLM remain frozen throughout all training phases to preserve the inherent foundational knowledge in both vision and language models. The projector and inner-adapter architecture require training to enhance multimodal capabilities. Our empirical investigations reveal that for the inner-adaptor architecture, applying a high learning rate can lead to overflow in training loss. To alleviate this issue, we devise a dual-stage pre-training procedure.}

\xie{In the first pre-training stage, the model configuration consists of only three components: the image encoder, the projector, and the large language model. The parameters of the image encoder and the large language model are frozen, while a high learning rate of $0.001$ is utilized to train a high-quality projector.}

\begin{table*}[t]
  \centering
  \renewcommand{\arraystretch}{1.3}
  \small 
  \scalebox{0.88}{
  \begin{tabular}{lccccccc}
  \toprule  

        Method &     Vision Encoder    &    Language Model &Data Scale &MME\textsuperscript{P} &   MMB-EN\textsuperscript{T}&  MMB-CN\textsuperscript{T}&   MMMU\textsuperscript{v}\\
         
         \midrule
  \multicolumn{7}{l}{\textit{Training with the LLM unfrozen}}\\
  \midrule
    mPLUG-Owl\cite{ye2023mplugowl}   &  CLIP-ViT-L  & Llama2 (7B)&1.1B& 967.3  &49.4 &-  & -\\
   Qwen-VL-Chat \cite{bai2023qwen}  &  CLIP-ViT-G  & Qwen (7B) &1.5B& 1487.6  &61.8 &56.3  & 37\\
   CogVLM  \cite{wang2023cogvlm} &  EVA2-CLIP-ViT-E  & Vicuna-v1.5 (7B) &1.5B& 1439.7  &65.8 &55.9  & 37.3\\

    mPLUG-Owl2 \cite{ye2024mplug}  &  CLIP-ViT-L  & Llama2 (7B) &400M& 1450.2  &66.0 & 60.3 &34.7 \\
   LLaVA-1.5\cite{llava}   &  CLIP-ViT-L  & Vicuna-v1.5 (7B) &1.2M& 1510.7  & 66.5&59.0  &35.7 \\
   LLaVA-1.5\cite{llava}   &  CLIP-ViT-L  & Vicuna-v1.5 (13B) &1.2M& 1531.3  & 69.2&65.0 &37.0 \\
    Honeybee \cite{cha2024honeybee}  &  CLIP-ViT-L & Vicuna-v1.5 (7B) &208M& 1584.2  &70.1& - &- \\
    Yi-VL \cite{ai2024yi}  &  CLIP-ViT-H  & Yi (6B) &125M& -  &68.4& 66.6 &39.1 \\
    DeepSeek-VL \cite{lu2024deepseek}  &  SAM-B and SigLIP-L  & DeepSeek (7B) &103M& -  &73.8& \textbf{71.4} &36.6 \\
    LLaVA-Llama3 \cite{llavallama3}  &  CLIP-ViT-L  &  Llama3 (8B) &1.2M& 1506.0  &68.9& 61.6 &36.8 \\
   \midrule
   \multicolumn{7}{l}{\textit{Training with the LLM frozen}}\\
   \midrule
   OpenFlamingov2  \cite{awadalla2023openflamingo} &   CLIP-ViT-L  & MPT (7B) &3B& -  &5.7 &14.4 & 28.8 \\
   Llama-AdapterV2\cite{gao2023llamaadav2}&  CLIP-ViT-L & Llama2 (7B) &0.6M& 972.7  &41.0 &-  & - \\
   MiniGPT-4 \cite{zhu2023minigpt}  &  EVA-CLIP-ViT-G  & Vicuna (13B) &5.1M& 866.6  & -& -&-  \\    
    BLIP-2  \cite{li2023blip} &  EVA-CLIP-ViT-G  & FlanT5XXL &129M& 1293.8  &- & - & - \\
    
    InstructBLIP \cite{InstructBLIP} &  EVA-CLIP-ViT-G  & Vicuna (13B) &130M& 1212.8  &44.0 & - &-  \\
    IAA-8\textsuperscript{$\dagger$}   &  CLIP-ViT-L  & Llama3 (8B) &1.2M& 1560.2  & 69.9 &64.2  &39.0 \\
    IAA-8   &  CLIP-ViT-L  & Llama3 (8B) &1.5M& 1581.8  & 72.7 &69.2 &39.8 \\
    IAA-14   &  CLIP-ViT-L  & Llama3 (8B) &1.5M& \textbf{1591.5}  & \textbf{74.9} &70.5 &\textbf{39.9} \\
     \bottomrule
  \end{tabular}
  }
  \caption{\xie{Results on general multimodal benchmarks}, where the data scale of 1.2M uniformly represents the data provided by LLaVA \cite{llava}. IAA-8\textsuperscript{$\dagger$} represents the model trained using 1.2M data.}
   \label{tab:eval}
\end{table*}


         
   

    

\begin{table}[t]
\centering
\renewcommand{\arraystretch}{1.3}
\small 
\scalebox{0.82}{
\begin{tabular}{lccccc}
\toprule  

    Method &     MMLU$\uparrow$ & C-Eval $\uparrow$&     BBH$\uparrow$ & Humaneval $\uparrow$& Math $\uparrow$\\
     
     \midrule

LLaVA-Llama3    &    55.8 & 40.5&    44.6 & 38.4& 12.3\\
IAA-8\textsuperscript{$\dagger$}    &  68.4  &51.3 &  52.8  &59.2&27.8\\

 \bottomrule
\end{tabular}
}
  \caption{Comparison on Text-only Benchmarks. IAA-8\textsuperscript{$\dagger$} denotes the model trained using the same 1.2M data as LLaVA-Llama3. IAA-8\textsuperscript{$\dagger$} is not impaired in terms of NLP ability, but LLaVA-Llama3 presents deteriorated results.}
   \label{tab:NLP}
\end{table}

\xie{In the second pre-training stage, the model architecture is expanded to incorporate the inner-adaptor for multimodal tasks. The training parameters now include both the projector and the newly integrated structures. The projector is initialized with the parameters derived from the preceding stage. For this stage, a lower learning rate of 2e-5 is adopted.}

\xie{Throughout the pre-training stages, the dataset employed consists of 558k image-text aligned pairs sourced from \cite{llava} and an additional 100K pairs from \cite{chen2024allava}. \cite{chen2024allava} provides a total of 664K image-text aligned data. We translate the first 100k pairs into Chinese and incorporated them into the training process to fortify the model's understanding of Chinese tasks. Over the course of these stages, we utilize a cumulative total of 658K data pairs.}

\subsubsection{\xie{Instruction Fine-tuning.}}
\xie{We perform instruction fine-tuning based on the model obtained from the second pre-training stage. Throughout this stage, the parameters of the large language model and the image encoder remain frozen. The dataset includes the fine-tuning dataset of 665K samples proposed by \cite{llava}, along with additional datasets including DocVQA (50K) \cite{docvqa}, VSR (10K) \cite{vsr}, ScienceQA (21K) \cite{siceqa}, and an in-house dataset (78.5K). Similar to the pre-training stage, we translate the first 40K entries of the 664K fine-tuning data proposed by \cite{chen2024allava} into Chinese and incorporate them into the instruction fine-tuning dataset. The aggregate quantity of data utilized in this stage amounts to 865K.}

\subsubsection{Grounding \xie{Fine-tuning.}}

\xie{Building upon the model fine-tuned with instructions, we further train a model specialized in visual grounding. The data used in this stage comprises RefCOCO \cite{refcoco}, COCO \cite{coco}, Flickr30k Entities \cite{flickr30k}, Objects365 \cite{shao2019objects365}, aggregating to approximately 2M data instances.}
These datasets improves the model’s capability of localizing fine-grained visual details. The inclusion of COCO and Objects365 assists the model in improving its ability to localize multiple targets.

\subsubsection{Implementation details.}

\xie{The detailed training information is summarized in Table \ref{tab:hpmsg}, mainly covering the hyperparameters used during the four-stage training process. The entire four-stage can be executed on a single node A800$\times$8 in 48 hours. All experiments utilize the zero technology provided by \cite{deepspeed} and the flash-attention v2 provided by \cite{dao2023flashattention}.}

\begin{figure*}[!htbp]
  \centering   \includegraphics[width=0.92\linewidth]{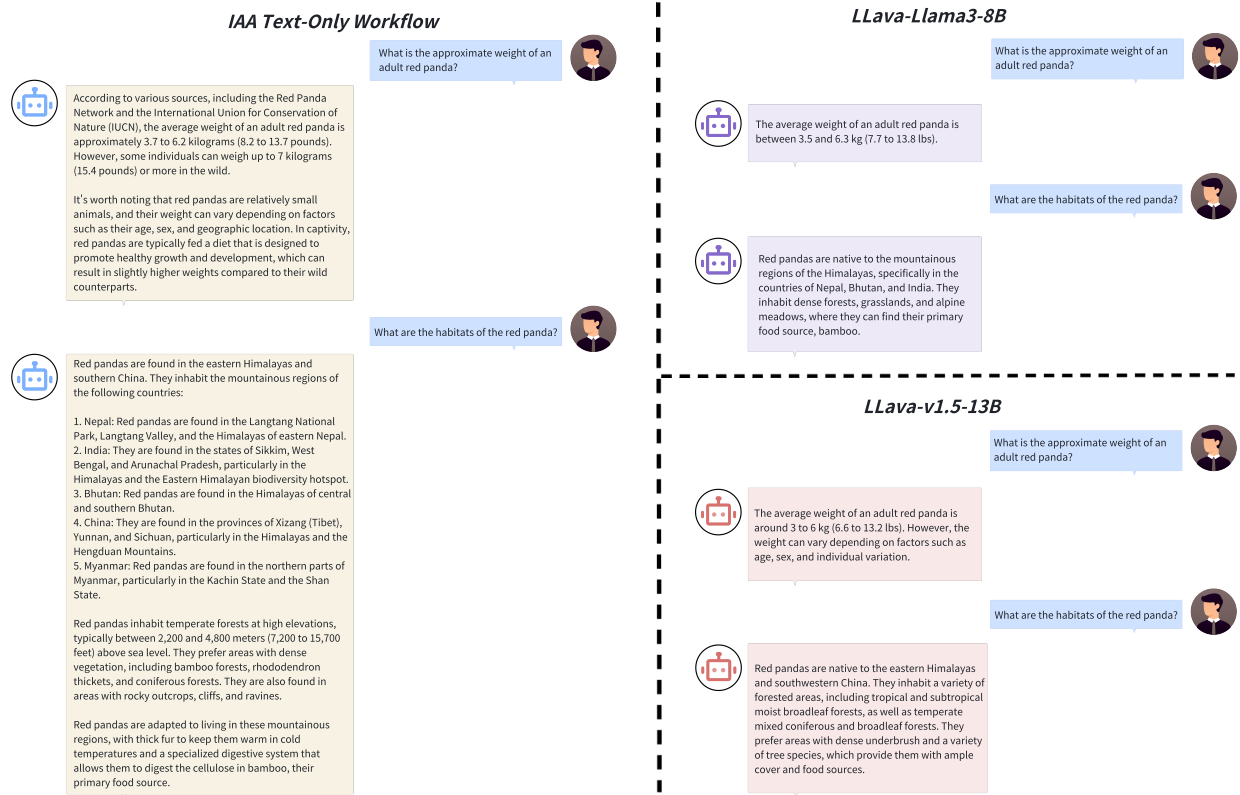}
    \caption{Comparison on text-only question answering.
}
    
    \label{fig:tflowc}
\end{figure*}

\begin{table*}[t]
  \centering
  \renewcommand{\arraystretch}{1.3}
  \small  
  \scalebox{1}{
  \begin{tabular}{cccccccccc}
  \toprule  

         
         \multirow{2}{*}{Method} & \multirow{2}{*}{Grounding Data Scale}&\multicolumn{3}{c}{RefCOCO} & \multicolumn{3}{c}{RefCOCO+}& \multicolumn{2}{c}{RefCOCOg}\\
         &&val&testA&testB&val&testA&testB&val&test \\
         \midrule
         KOSMOS-2 \cite{peng2023kosmos} & 20M & 52.3&57.4&47.3&45.5&50.7&42.2&60.6&61.7 \\
         OFA-L \cite{wang2022ofa} & 10M & 80.0&83.7&76.4&68.3&76.0&61.8&67.6 &67.6\\
         Shikra \cite{chen2023shikra} & 4M & 87.0&90.6&80.2&81.6&87.4&72.1&82.3&82.2 \\
         MiniGPT-v2 \cite{chen2023minigpt} & $\sim$21M & 88.7&91.7&85.3&80.0&85.1&74.5&84.4&84.7 \\
         Ferret \cite{you2023ferret} & 8.7M & 87.5&91.4&82.5&80.1&87.4&73.1&83.9&84.8 \\
         PINK \cite{xuan2024pink} & 5M &88.7&92.1&84.0&81.8&88.2&73.9&83.9&84.3 \\
         IAA-8 & 2M&89.2 &92.6&83.7&82.1 &88.6 &73.7 &84.4 &84.7  \\
         IAA-14 & 2M& \textbf{90.2}&\textbf{92.9}&\textbf{85.4}&\textbf{83.4} & \textbf{89.0}&\textbf{76.7} &\textbf{85.0} &\textbf{85.1 } \\
         

     \bottomrule
  \end{tabular}}
  \caption{\xie{Comparisons on visual grounding benchmarks. Our approach achieves competitive performance trained on relatively limited datasets.}}
   \label{tab:grounding_c}
\end{table*}

\subsection{\xie{Experimental Results}}
\subsubsection{\xie{Main Results on General Multimodal Benchmarks.}}
\xie{To assess the multimodal capabilities of our approach, we employ widely recognized benchmarks that are closely related to multimodal tasks: MME\textsuperscript{P} \cite{mme}, MMBench-EN\textsuperscript{T} \cite{liu2023mmbench}, MMBench-CN\textsuperscript{T} \cite{liu2023mmbench}, and MMMU\textsuperscript{v} \cite{yue2024mmmu}. These benchmarks are renowned for presenting significant challenges across a diverse range of practical tasks. For evaluation purposes, we adhere to a zero-shot testing protocol, a strict methodology that tests models on unseen data without additional training. Moreover, we categorize comparative methods into two distinct categories: those trained with a frozen language model and those trained with an unfrozen language model. To provide a comprehensive analysis, we show the scale of the data utilized for each method, along with the variations in the image encoders employed. Detailed results of our evaluations are tabulated in Table \ref{tab:eval}. To ensure a fair and equitable comparison, we choose methods that leverage a base language model with a comparable parameter scale, and the reported metrics for competing methods are based solely on officially published data, avoiding any local testing results.}

\xie{Owing to the inherent strengths of our proposed architecture, our method exhibits substantial superiority over those trained with frozen language model. As the current mainstream approach, models trained with unfrozen language models typically achieve better multimodal performance, albeit at the cost of diminished NLP capabilities. We list several state-of-the-art methods adhering to this training paradigm. Compared to Honeybee \cite{cha2024honeybee}, Yi-VL \cite{ai2024yi}, and Deepseek-VL \cite{lu2024deepseek}, our method achieves competitive or even superior performance on certain metrics, with an extremely small training data scale. Using the same data scale of 1.2 million, IAA-8 outperforms LLaVA-Llama3. Additionally, IAA-14 with 14 insertion layers achieves better results than IAA-8 with an 8-layer configuration. 
\todo{Furthermore, we compare our approach with LLaVA-Llama3 \cite{llavallama3} on NLP benchmarks, including MMLU and C-Eval. The results of NLP benchmarks are summarized in Table \ref{tab:NLP}. Our language model is not impaired in terms of NLP ability, but LLaVA-Llama3 trained on the same data shows deteriorated results on both MMLU and C-Eval. Our method surpasses LLaVA-Llama3 across all metrics, indicating that our architecture is superior to the mainstream LLaVA architecture.
The performance of various models on the plain text dialog task is illustrated in Figure 3. It is evident that the text-only workflow of the Inner-Adaptor Architecture (IAA) preserves the original conversational capabilities of the language model. In contrast, open-source multimodal large language models such as LLaVA-Llama3 and LLaVA-v1.5 are more impacted by multimodal data. When queried with the same question, LLaVA-Llama3 and LLaVA-v1.5 produce notably shorter responses. This is directly related to the fact that a large amount of the multimodal training data has shorter text lengths. Fine-tuning the large language model affects its ability to fully understand content and generate more comprehensive responses.
}}

\subsubsection{\xie{Results on Visual Grounding Benchmarks.}}


\xie{To evaluate the effectiveness of our model in the visual grounding task, we perform evaluations utilizing the widely accepted benchmarks RefCOCO \cite{refcoco}, RefCOCO+ \cite{refcocop}, and RefCOCOg \cite{refcocog}, with the corresponding results illustrated in Table \ref{tab:grounding_c}. The methods for comparison are all models trained for the grounding task under an auto-regressive strategy. The results reveal that our method is capable of achieving competitive performance, even when trained on relatively limited datasets. In our analysis, to ensure fairness, we exclude models trained on extremely  large-scale datasets, such as CogVLM-grounding \cite{wang2023cogvlm} with 1.5B image-text pairs and 40M grounding data, as well as those leveraging pre-trained object detection models, exemplified by LLaVA-Grounding \cite{llavag} and Groma \cite{ma2024groma}. }


\subsubsection{\xie{Efficiency in Deployment.}}
\xie{Currently, high-performance multimodal models typically require the unfreezing of the large language model for training. CogVLM \cite{wang2023cogvlm} highlights the substantial difficulty in developing a model that excels in both multimodal comprehension and visual grounding tasks simultaneously. To address this, it adopts a dual-model strategy, specifically training one model for general multimodal capabilities and another for visual grounding abilities. In this context, deploying a high-quality language model, a multimodal model with outstanding general performance, and a model endowed with proficient visual grounding skills concurrently on a single GPU would demand an estimated 50GB of memory. Our proposed approach, facilitated by the inner-adaptor architecture, ingeniously combines superior general multimodal competencies and robust visual grounding capacities, while concurrently safeguarding the inherent prowess of the original large language model. Specifically, with an 8-layer inner-adaptor configuration, our model exhibits a significantly reduced memory footprint, hovering around 30GB.}

\subsection{Ablation Study
}
\begin{table*}[t]
  \centering
  \small              
  \begin{tabular}{cccccccccc}
  \toprule  

         
         \multirow{2}{*}{Model architecture} & \multicolumn{5}{c}{Trainable modules} & \multirow{2}{*}{MME\textsuperscript{P}}&
         \multirow{2}{*}{MMB-EN\textsuperscript{T}}&
         \multirow{2}{*}{MMB-CN\textsuperscript{T}}&
         \multirow{2}{*}{MMMU\textsuperscript{v}}\\
&Projector&I-Layers(8)&$EL_{mm}$&$LH_{mm}$&Zero-Gates&&&&\\
\midrule

    Figure \ref{fig:subarch}(a)   & $\checkmark$ & $\checkmark$ & $\checkmark$ &$\checkmark$ &$\checkmark$ &1425.4 &72.4&65.0&38.2\\
    Figure \ref{fig:subarch}(b)   & $\checkmark$ & $\checkmark$ & $\checkmark$ &$\checkmark$ &$\checkmark$ &1556.0&72.7&68.5&39.6  \\
    Figure \ref{fig:subarch}(c)   & $\checkmark$ & $\checkmark$ & $\times$ &$\times$ &$\times$ &1563.4&72.6&68.7&39.6\\
    Figure \ref{fig:subarch}(c)   & $\checkmark$ & $\checkmark$ & $\checkmark$ &$\checkmark$ &$\times$ & 1581.8  & 72.7 &69.2 &39.8 \\
     \bottomrule
  \end{tabular}
  \caption{Ablation study for the exploration of inner-adaptor related structures.}
   \label{tab:aba_strc}
\end{table*}

\begin{table}[t]
  \centering
  \small              
  \begin{tabular}{ccccc}
  \toprule  

         
         \multicolumn{3}{c}{Training Stages} & \multirow{2}{*}{MME\textsuperscript{P}}&\multirow{2}{*}{MMMU\textsuperscript{v}}\\
Satge1-P&Satge2-P&Instruction-F &&\\
\midrule
    $\times$ & $\checkmark$ & $\checkmark$ & 1512.1   &39.3\\
  $\checkmark$ & $\times$ & $\checkmark$ & 1565.4&39.5 \\
   $\checkmark$ & $\checkmark$ & $\checkmark$ & 1581.8   &39.8\\

     \bottomrule
  \end{tabular}
  \caption{Comparison of the training stages.}
   \label{tab:train_stage}
\end{table}

\begin{table}[t]
  \centering
  \small              
  \begin{tabular}{cccc}
  \toprule  

        Number of I-Layers &     
     MME\textsuperscript{P} &    MMB-EN\textsuperscript{T}&  MMB-CN\textsuperscript{T}\\
     \midrule
         8 &1581.8   &72.7&69.2\\
        14&1591.5   &74.9&70.5\\
        22&1531.7&76.0&70.4\\

     \bottomrule
  \end{tabular}
  \caption{Ablations on the number of insertion layers.}
   \label{tab:layer}
\end{table}

\begin{table}[t]
  \centering
  \small   
  \scalebox{0.9}{
  \begin{tabular}{cccc}
  \toprule  

         &     
     MME\textsuperscript{P} &MMB-EN\textsuperscript{T}&  MMB-CN\textsuperscript{T}\\
     \midrule
     LLaVA-Llama3 (1.2M)&1506.0  &68.9& 61.6 \\
         IAA-8 (1.2M) &1560.2 &69.9&64.2 \\
        IAA-8 (1.5M)&1581.8 &72.7&69.2  \\
  
     \bottomrule
  \end{tabular}
  }
  \caption{The impact of the training data.}
   \label{tab:datacompare}
\end{table}

\subsubsection{\xie{Structure Analysis.}}
\xie{In the exploration of the structure, we furnish quantitative results for validation in Table \ref{tab:aba_strc}. With an 8-layer insertion scheme as our baseline configuration, we observe that incremental architectural enhancements consistently improve performance metrics across the board. Specifically, the comparison between rows 1, 2, and 4 highlights the benefits of architectural refinement. Moreover, the contrast between rows 3 and 4 demonstrates that the integration of a specialized embedding layer and language model head for multimodal data processing significantly boosts performance.}

\subsubsection{Comparison of Training Stages.}
\xie{Through empirical evidence detailed in Table \ref{tab:train_stage}, we validate the effectiveness of our two-stage pre-training methodology. It can be observed that the model lacking the first stage of alignment training exhibits notably poorer performance. When the projector and insertion layers are engaged in joint pre-training, it is essential to maintain a learning rate of approximately 2e-5 to prevent loss overflow. However, this strategy leads to suboptimal alignment training for the projector, which negatively affects the model's final performance. }
\leng{
Furthermore, although the model performs adequately when skipping the second pre-training stage, it ultimately fails to replicate the outstanding results achievable through the complete two-stage pre-training process. This disparity emphasizes the critical significance of the additional pre-training stage in enhancing the model's overall effectiveness.
}

\subsubsection{Impact of Insertion Layer Quantities.}
\xie{We explore the effect of varying numbers of insertion layers, which are presented in Table \ref{tab:layer}. The experimental results indicate that increasing the number of insertion layers from 8 to 14 yields enhancements in all performance metrics. However, it is imperative to acknowledge that an increase in insertion layers simultaneously impacts the model's efficiency. We advocate that an 8-layer configuration is adequate to effectively address foundational requirements.}

\subsubsection{Training Data Influence Assessment.}
\xie{To delineate the impact of data on model performance, we present comparative results in Table \ref{tab:datacompare}. The baseline, outlined in the first row, showcases the performance of LLaVA-Llama3 \cite{llavallama3} utilizing the LLaVA architecture and the 1.2 million dataset provided by \cite{llava}. Subsequent experimentation, as delineated in the second row, emphasizes the pronounced superiority of our proposed architecture over LLaVA. Additionally, we enrich the training corpus with an extra 0.3 million records, mainly encompassing Chinese data. As a result, our model achieves substantial improvements in all metrics, especially on the Chinese evaluation set MMBench-CN\textsuperscript{T}.}

\wang{
\subsubsection{Limitations}
The method of extending multimodal capabilities by freezing the language model will introduce certain additional parameters. Compared to the approach of training with an \xie{unfrozen} language model, the inference speed of the model will be reduced. To mitigate this issue, we \xie{extend} the key-value cache mechanism to the insertion layers. Based on the MME dataset, compared to the LLaVA architecture, the average inference time of our 8-layer structure increases from 0.103s to 0.124s, which we deem to be within a relatively reasonable range.}

\section{Conclusion}

\xie{In this paper, we introduce the Inner-Adaptor Architecture, which is designed to enhance the general multimodal and visual grounding capabilities of LLMs. Through a series of architectural exploration experiments, we demonstrate that training with a frozen language model can surpass the multimodal performance of the models with fine-tuned LLMs. Our proposed model has achieved state-of-the-art performance across a multitude of publicly available evaluation datasets. Moreover, after deployment, our approach incorporates dual workflows, thereby preserving the NLP proficiency of the language model. The flexibility of the Inner-Adaptor Architecture provides the potential for extension to additional modalities, which is a direction for future exploration.}

\bibliography{aaai25}

\appendix

\newpage
\section{The Details of Dateset}

In this section, we introduce the datasets IAA uses at different stages, along with the possible download links for these datasets in detail. 

\subsection{Pre-training}

Throughout the pre-training stages, the dataset employed consists of 558k image-text aligned pairs sourced from LLaVA and an additional 100K pairs from ALLaVA. ALLaVA provides a total of 664K image-text aligned data. We translate the first 100k pairs into Chinese and incorporated them into the training process to fortify the model's understanding of Chinese tasks. Over the course of these stages, we utilize a cumulative total of 658K data pairs.

\begin{links}
\link{558k pairs from LLaVA}{https://huggingface.co/datasets/liuhaotian/LLaVA-Pretrain}
\link{ALLaVA}{https://huggingface.co/datasets/FreedomIntelligence/ALLaVA-4V}
\end{links}

\subsection{Instruction Fine-tuning}

We perform instruction fine-tuning based on the model obtained from the second pre-training stage. Throughout this stage, the parameters of the large language model and the image encoder remain frozen. The dataset includes the fine-tuning dataset of 665K samples proposed by LLaVA, along with additional datasets including DocVQA (50K), VSR (10K), ScienceQA (21K), and an in-house dataset (78.5K). Similar to the pre-training stage, we translate the first 40K entries of the 664K fine-tuning data proposed by ALLaVA into Chinese and incorporate them into the instruction fine-tuning dataset. The aggregate quantity of data utilized in this stage amounts to 865K.

\begin{links}
\link{665K samples from LLaVA}{https://huggingface.co/datasets/liuhaotian/LLaVA-Instruct-150K}
\link{DocVQA (50K)}{https://huggingface.co/datasets/cmarkea/doc-vqa}
\link{VSR (10K)}{https://github.com/cambridgeltl/visual-spatial-reasoning/}
\link{ScienceQA (21K)}{https://github.com/lupantech/ScienceQA}
\link{ALLava}{https://huggingface.co/datasets/FreedomIntelligence/ALLaVA-4V}
\end{links}
\subsection{Grounding Fine-tuning}

Building upon the model fine-tuned with instructions, we further train a model specialized in visual grounding. The data used in this stage comprises RefCOCO, COCO, Flickr30k Entities, Objects365, aggregating to approximately 2M data instances.
These datasets improves the model’s capability of localizing fine-grained visual details. The inclusion of COCO and Objects365 assists the model in improving its ability to localize multiple targets.

\begin{links}
\link{RefCOCO}{https://github.com/lichengunc/refer}
\link{COCO}{https://cocodataset.org/}
\link{Flickr30k Entities}{https://github.com/BryanPlummer/flickr30k_entities}
\link{Objects365}{https://www.objects365.org/}
\end{links}

\section{Supplementary Display}

\subsubsection{Multimodal Capability.}
Figures \ref{fig:s1} and \ref{fig:s2} showcase the capabilities of the Inner-Adaptor Architecture (IAA) in encyclopedia question answering, image comprehension, text recognition, and writing.

\subsubsection{Grounding Capability.}
Figure \ref{fig:s3} presents the multi-object detection capability of IAA, while Figure \ref{fig:s4} demonstrates its detection capability for fine-grained perception.

    

\begin{figure*}[!htbp]
  \centering   \includegraphics[width=1.0\linewidth]{{sample1_2}}
    \caption{Samples of image comprehension and general knowledge question answering.}
    
    \label{fig:s1}
\end{figure*}

\begin{figure*}[!htbp]
  \centering   \includegraphics[width=0.9\linewidth]{{sample3_4}}
    \caption{Samples of text recognition and writing ability.}
    
    \label{fig:s2}
\end{figure*}

\begin{figure*}[!htbp]
  \centering   \includegraphics[width=0.8\linewidth]{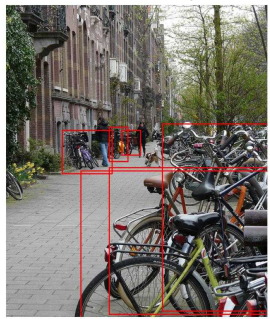}
    \caption{Samples of multi-object detection
.}
    
    \label{fig:s3}
\end{figure*}

\begin{figure*}[!htbp]
  \centering   \includegraphics[width=0.8\linewidth]{{g_sample34}}
    \caption{Samples of fine-grained detection.}
    
    \label{fig:s4}
\end{figure*}

\end{document}